\documentclass{article}

\usepackage{microtype}
\usepackage{graphicx}
\usepackage{subfigure}
\usepackage{booktabs} % for professional tables
\usepackage{amsmath}
\usepackage{tikz}

\usepackage[colorinlistoftodos]{todonotes}

\DeclareMathOperator*{\argmin}{argmin}

\usepackage{hyperref}

\usepackage[accepted]{icml2024}

\usepackage{amsmath}
\usepackage{amssymb}
\usepackage{mathtools}
\usepackage{amsthm}
\usepackage{multirow}

\usepackage[capitalize,noabbrev]{cleveref}

\theoremstyle{plain}

\theoremstyle{definition}

\theoremstyle{remark}

\icmltitlerunning{Piecewise Normlizing Flows}

\begin{document}

\twocolumn[
\icmltitle{Piecewise Normalizing Flows}

% It is OKAY to include author information, even for blind
% submissions: the style file will automatically remove it for you
% unless you've provided the [accepted] option to the icml2023
% package.

% List of affiliations: The first argument should be a (short)
% identifier you will use later to specify author affiliations
% Academic affiliations should list Department, University, City, Region, Country
% Industry affiliations should list Company, City, Region, Country

% You can specify symbols, otherwise they are numbered in order.
% Ideally, you should not use this facility. Affiliations will be numbered
% in order of appearance and this is the preferred way.
\icmlsetsymbol{equal}{*}

\begin{icmlauthorlist}
\icmlauthor{Harry Thomas Jones Bevins}{kav,cav}
\icmlauthor{Will Handley}{kav,cav}
\icmlauthor{Thomas Gessey-Jones}{kav,cav}

\end{icmlauthorlist}

\icmlaffiliation{cav}{Cavendish Astrophysics,
  University of Cambridge,
  Cavendish Laboratory,
  JJ Thomson Avenue,
  Cambridge, CB3 0HE}
\icmlaffiliation{kav}{Kavli Institute for Cosmology,
  University of Cambridge,
  Madingley Road,
  Cambridge, CB3 0HA}

\icmlcorrespondingauthor{Harry Thomas Jones Bevins}{htjb2@cam.ac.uk}
\icmlcorrespondingauthor{Will Handley}{wh260@cam.ac.uk}
\icmlcorrespondingauthor{Thomas Gessey-Jones}{tg400@cam.ac.uk}

% You may provide any keywords that you
% find helpful for describing your paper; these are used to populate
% the "keywords" metadata in the PDF but will not be shown in the document
\icmlkeywords{Machine Learning, ICML}

\vskip 0.3in
]

% this must go after the closing bracket ] following \twocolumn[ ...

% This command actually creates the footnote in the first column
% listing the affiliations and the copyright notice.
% The command takes one argument, which is text to display at the start of the footnote.
% The \icmlEqualContribution command is standard text for equal contribution.
% Remove it (just {}) if you do not need this facility.

%\printAffiliationsAndNotice{}  % leave blank if no need to mention equal contribution
\printAffiliationsAndNotice{} % otherwise use the standard text.

\begin{abstract}
Normalizing flows are an established approach for modelling complex probability densities through invertible transformations from a base distribution. However, the accuracy with which the target distribution can be captured by the normalizing flow is strongly influenced by the topology of the base distribution. A mismatch between the topology of the target and the base can result in a poor performance, as is typically the case for multi-modal problems. A number of different works have attempted to modify the topology of the base distribution to better match the target, either through the use of Gaussian Mixture Models \citep{izmailov2020semi, ardizzone2020training, hagemann2021stabilizing} or learned accept/reject sampling \citep{Stimper2021}. We introduce piecewise normalizing flows which divide the target distribution into clusters, with topologies that better match the standard normal base distribution, and train a series of flows to model complex multi-modal targets.
We demonstrate the performance of the piecewise flows using some standard benchmarks and compare the accuracy of the flows to the approach taken in \citet{Stimper2021} for modelling multi-modal distributions. We find that our approach consistently outperforms the approach in \citet{Stimper2021} with a higher emulation accuracy on the standard benchmarks.
\end{abstract}

\section{Introduction}

Normalizing flows~(NF) are an effective tool for emulating probability distributions given a set of target samples. They are bijective and differentiable transformations between a base distribution, often a standard normal distribution, and the target distribution \citep{Papmakarios_NF_review_2019}. They have been applied successfully in a number of fields such as the approximation of Boltzmann distributions \citep{Wirnsberger2020}, cosmology \citep{Alsing2021, Bevins2022b, Friedman2022}, image generation \citep{Ho2019,Grcic2021}, audio synthesis \citep{Oord2018}, density estimation \citep{MAFs,Huang2018} and variational inference \citep{Rezende2015} among others.

\begin{figure*}[t!]
    \centering
    \includegraphics[width=0.6\linewidth]{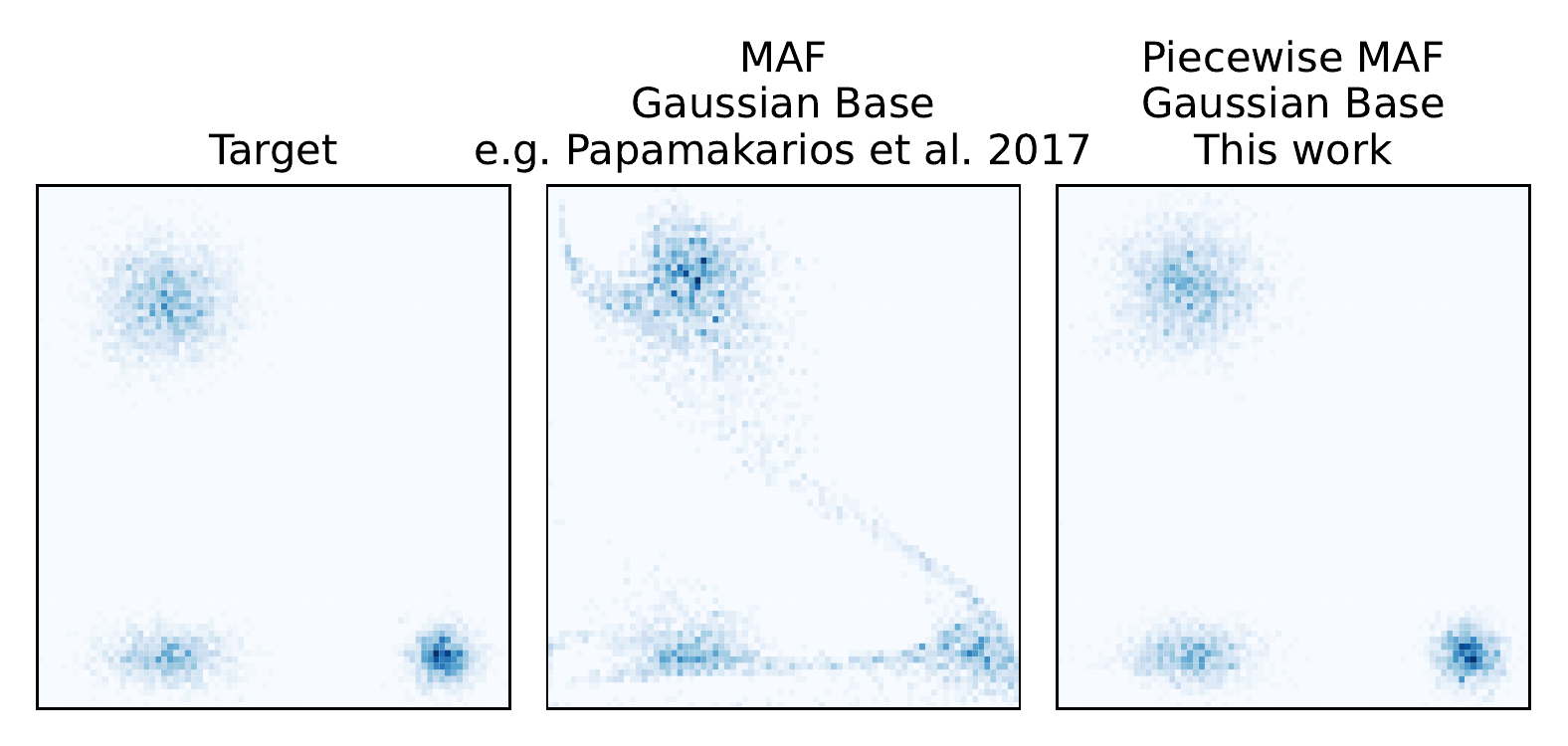}
    \caption{A simple demonstration of the piecewise normalizing flow~(NF) described in this paper and a single NF trained on the same multimodal target distribution. We use early stopping to train the different normalizing flows. For the single NF we see bridging between the different clusters that is not present in the piecewise approach. The architecture of the MAF and piecewise MAF components are chosen such that the two have approximately the same number of hyperparameters.}
    \label{fig:example}
\end{figure*}

When the target distribution is multi-modal, NFs can take a long time to train well, and often feature `bridges' between the different modes that are not present in the training data (see \cref{fig:example}). These bridges arise due to differences between the topology of the base distribution and the often complex target distribution \citep{Cornish2019RelaxingBC} and is a result of the homeomorphic nature of the NFs \citep{Runde2007}.

One approach to deal with multi-modal distributions is to modify the topology of the base distribution. This was demonstrated in \cite{Stimper2021}, in which the authors developed a base distribution derived using learned rejection sampling. %Typically, the bijective nature of the normalizing flows is lost when new tools are developed to learn multi-modal probability distributions, however this was not the case for the resampled base distribution approach discussed in \cite{Stimper2021}.
There are a number of other approaches to solve the mismatch between base and target distribution topologies such as; continuously indexing flows \citep{Cornish2019RelaxingBC}, augmenting the space the model fits in \citep{Huang2018} and adding stochastic layers \citep{wu2020stochastic} among others.

A few works have attempted to use Gaussian mixture models for the base distribution \citep[e.g.][]{MAFs, izmailov2020semi} and have extended various NF structures to perform classification while training. For example, \citet{izmailov2020semi} showed that a high degree of accuracy can be accomplished if data is classified in modes in the latent space based on a handful of labelled samples. \citet{ardizzone2020training} uses the information bottleneck to perform classification via a Gaussian mixture model. Similar ideas were explored in \citet{hagemann2021stabilizing}. Recently, it was shown that the boundaries between different modes could be effectively learnt during training using Voronoi clustering \cite{chen2022}.

\textbf{Our main contribution} is to classify samples from our target distribution using a clustering algorithm \emph{before} training and train normalizing flows on the resultant clusters in a piecewise approach where each piece constitutes a NF with a Gaussian base distribution. This improves the accuracy of the flow and reduces the computational resources required for accurate training for two reasons. Firstly, it is parallelizable and each piece can be trained simultaneously, and secondly if sufficient clusters are used, the topology of each piece of the target distributions is closer to that of the Gaussian base distribution. We compare the accuracy of our results against the approach put forward in \cite{Stimper2021} using a series of benchmark multi-modal distributions. %As far as we are aware, this is the first attempt at such a piecewise normalizing flow.

\section{Normalizing Flows}
\label{sec:normalizing_flows}

If we have a target distribution $X$ and a base distribution $Z\sim\mathcal{N}(\mu=0, \sigma=1)$ then via a change of variables we can write
\begin{equation}
    p_X(x) = p_Z(z)\bigg|\mathrm{det}\bigg(\frac{\partial z}{\partial x}\bigg)\bigg|,
\end{equation}
where the base distribution is chosen to be Gaussian so that it is easy to sample and evaluate.
If there is a bijective transformation between the two $f$ then $z$ is given by $f^{-1}_\theta(x)$ and the change of variable formula becomes
\begin{equation}
    p_X(x) = p_Z(f^{-1}_\theta(x))\bigg|\mathrm{det}\bigg(\frac{\partial f^{-1}_\theta(x)}{\partial x}\bigg)\bigg|,
    \label{eq:change_of_variables}
\end{equation}
where $\theta$ are the parameters describing the transformation from the base to target distribution. Here, the magnitude of the determinant of the Jacobian represents the volume contraction from $p_Z$ to $p_X$. The transformation $f$ corresponds to the invertible normalizing flow between $p_X$ and $p_Z$ and $\theta$ to the hyperparameters of the flow. In our work we use Masked Autoregressive Flows \citep[MAFs,][]{MAFs} which are based on the Masked Autoencoder for Distribution Estimation \citep[MADEs, ][]{MADEs}.

For a MAF, the multidimensional target distribution is decomposed into a series of one dimensional conditional distributions
\begin{equation}
p_X(x) = \prod_i p(x_i| x_1, x_2, ..., x_{i-1}),
\end{equation}
where the index $i$ represents the dimension. Each conditional probability is modelled as a Gaussian distribution with some standard deviation, $\sigma_i$, and mean, $\mu_i$, which can be written as
\begin{equation}
    \begin{aligned}
    P(x_i| x_1, &x_2, ..., x_{i-1}) = \\ & \mathcal{N}(\mu_i(x_1, x_2, ..., x_{i-1}), \sigma_i(x_1, x_2, ..., x_{i-1})).
    \end{aligned}
\end{equation}
$\sigma_i$ and $\mu_i$ are functions of $\theta$ and are output from the network. The optimum values of the network weights and biases are arrived at by training the network to minimize the Kullback-Leibler~(KL) divergence between the target distribution and the reconstructed distribution
\begin{equation}
    \mathcal{D} (p_X(x)||p_{\theta}(x)) = -\mathbb{E}_{p_X(x)}[\log p_\theta(x)] + \mathbb{E}_{p_X(x)}[\log p_X(x)].
\end{equation}
The second term in the KL divergence is independent of the network and so for a minimization problem we can ignore this and since
\begin{equation}
    -\mathbb{E}_{p_X(x)}[\log p_\theta(x)] = -\frac{1}{N} \sum_{j=0}^N \log p_\theta(x_j),
\end{equation}
where $j$ corresponds to a sample in the set of $N$ samples, our minimization problem becomes
\begin{equation}
    \argmin_\theta -\sum^N_{j=0} \log p_\theta(x_j).
\end{equation}
In practice the flows are trained to transform samples from the target distribution into samples from the base distribution and the loss is evaluated by a change of variables
\begin{equation}
    \argmin_\theta -\sum_{j=0}^N [\log p_Z(z_j^\prime) + \log \bigg|\frac{\partial z_j^\prime}{\partial x_j}\bigg|],
    \label{eq:actual_loss}
\end{equation}
where $z^\prime$ is a function of $\theta$ \citep{Alsing2019, Stimper2021}. See Fig. 1 of \citet{Bevins2022a} for an illustration of the MADE network.
Typically, many MADE networks are chained together to improve the expressivity of the MAF \citep{MAFs}.

\section{Clustering and Normalizing Flows}

\begin{figure*}
    \centering
    \includegraphics[width=0.8\linewidth]{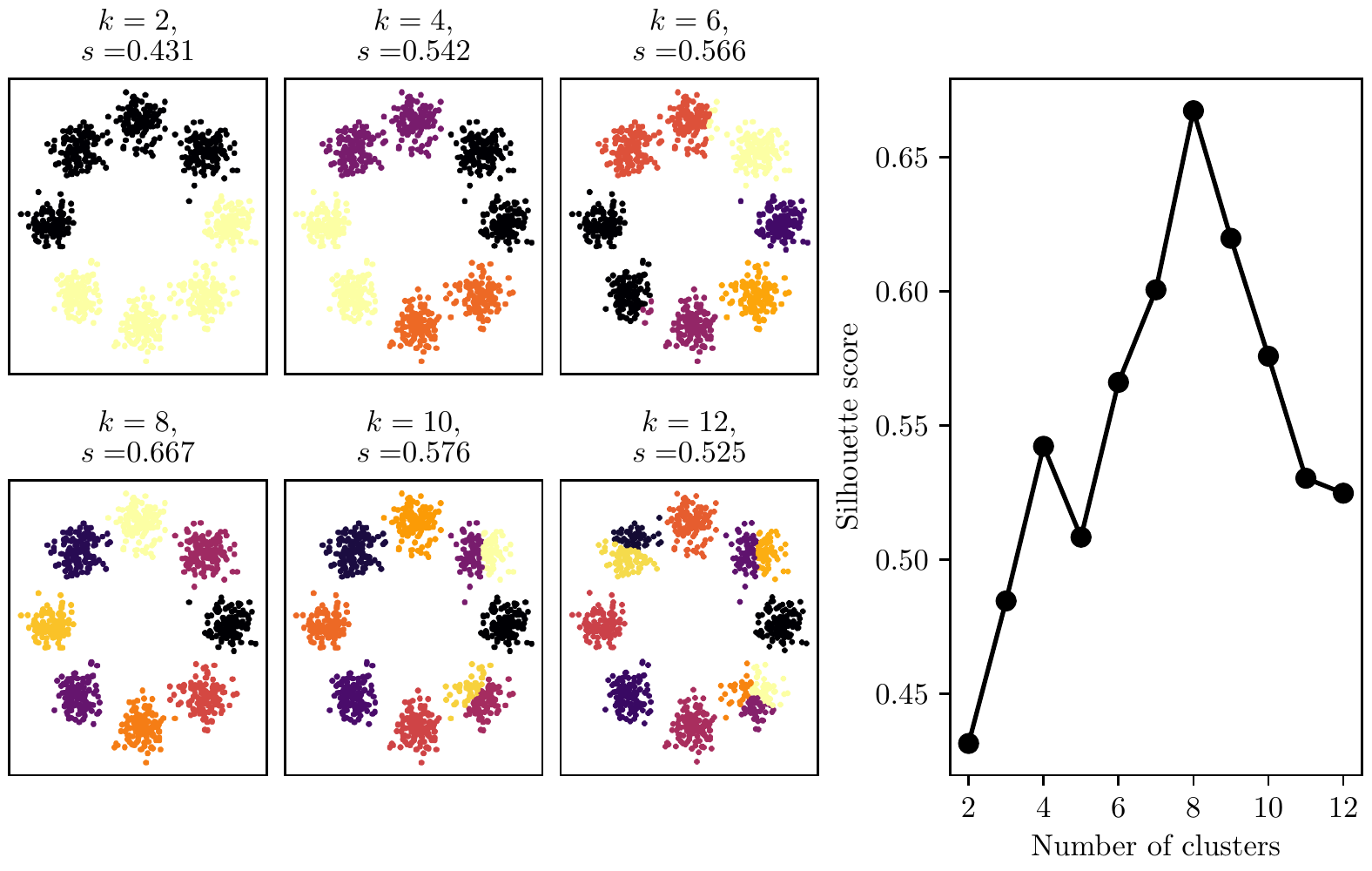}
    \caption{The figure shows how the silhouette score, $s$, can be used to determine the number of clusters needed to describe the data. In the example $s$ peaks for eight clusters corresponding to one for each Gaussian in the circle.}
    \label{fig:silhouette}
\end{figure*}

We propose a piecewise approach to learning complex multi-modal distributions, in which we decompose the target distribution into clusters before training and train individual MAFs on each cluster. For $k$ clusters of samples $x_k$ we therefore have a set
\begin{equation}
    x = \{x_1, x_2, ..., x_k\}~~\textnormal{and}~~p_{\theta} = \{p_{\theta_1},p_{\theta_2}, ..., p_{\theta_k}\}.
\end{equation}

Typically, if our samples for our target distribution have been drawn with some sample weights, $w_j$, (i.e. using an MCMC \citep[e.g.][]{Foreman-Mackey2013} or Nested Sampling \citep{Skilling2006, Handley2015} algorithm) %we write our loss function over $N$ samples as
%\begin{equation}
%    \argmin_\theta -\sum^N_{j=0} w_j \log p_\theta(x_j).
%\end{equation}
then each set of samples in each cluster has associated weights
\begin{equation}
    w = \{w_1, w_2, ..., w_k\}.
\end{equation}
and our total loss over $N_\mathrm{clusters}$ is given by
\begin{equation}
    \begin{aligned}
    &\argmin_\theta -\bigg[\sum^{N_0}_{j=0} w_{0, j} \log p_{\theta_1}(x_{0, j}) \\ & + \sum^{N_1}_{j=0} w_{1, j}\log p_{\theta_2}(x_{1, j}) + ...  +\sum^{N_k}_{j=0} w_{k, j}\log p_{\theta_k}(x_{k, j})\bigg],
    \end{aligned}
\end{equation}
which can be written as
\begin{equation}
    \argmin_\theta -\sum^{N_\mathrm{cluster}}_{k=0} \sum^{N_k}_{j=0} w_{k, j}\log p_{\theta_k}(x_{k, j}).
    \label{eq:loss}
\end{equation}

Hence, the loss function for the entire network is just the linear sum of the losses for individual cluster MAFs. We thus train each cluster's MAF separately, using an ADAM optimizer \citep{Kingma2014AdamAM}. Since each cluster constitutes a simpler target distribution than the whole, we find that we need fewer computing resources when training a piecewise NF in comparison to training a single NF on the same problem (see \cref{sec:benchmarking} and \cref{tab:benchmark_times}). One can train each piece of the PNF in parallel, leading to further (wall time) computational gains. %Further, each network in the piecewise ensemble can be trained in parallel, and this can significantly speed up training.

To determine the number of clusters needed to best describe the target distribution, we use an average silhouette score, where the silhouette score for an individual sample is a measure of how similar a sample is with the samples in its cluster compared to samples outside its cluster \citep{Rousseeuw1987}. The score ranges between -1 and 1 and peaks when the clusters are clearly separated and well-defined, as in \cref{fig:silhouette}.

\begin{figure*}
    \centering
    \begin{tikzpicture}[rednode/.style={rectangle, draw=red!60, fill=red!5, very thick, minimum size=1.5mm},
                bluenode/.style={circle, draw=blue!60, fill=blue!5, very thick, minimum size=5mm},
                greennode/.style={circle, draw=green!60, fill=green!5, very thick, minimum size=5mm},
                node distance=0.5cm and 2cm,
                remember picture]

        \node[bluenode, text width=1.5cm, align=center](layer1_center) at (-7, 0) {$x\sim p(x)$};

        \node[rectangle, very thick, draw=pink!80, text=black, minimum width = 9cm, minimum height=7.5cm, fill=pink!20] (r) at (-1,0) {};
        \node[text width=9cm, font=\Large] at (0.3, 3.25) {Piecewise Normalizing Flows};
        
        \node[rednode, text width=1.5cm, align=center](classifier) at (-4, 0) {Classifier};

        \node[rednode, text width=1.5cm, align=center, minimum width=2.5cm](maf1) at (0, 2.5) {MAF$_{k=1}$};
        
        \node[rednode, text width=1.5cm, align=center, minimum width=2.5cm](maf2) at (0, 0.75) {MAF$_{k=2}$};

        \node[rednode, text width=1.5cm, align=center, minimum width=2.5cm](maf3) at (0, -1) {MAF$_{k=3}$};

        \node[rednode, text width=1.5cm, align=center, minimum width=2.5cm](maf4) at (0, -2.75) {MAF$_{k=4}$};

        \node[greennode, text width=1.5cm, align=center](output_1) at (6, 0) {$x \sim p_\theta (x)$};

        \draw[->, very thick](layer1_center.east) -- node[rectangle, very thick, draw=black, fill=gray!20] {$N$} (classifier);
        \draw[->, very thick](classifier.east) --  node[rectangle, very thick, draw=black, fill=gray!20] {$N_1$} (maf1.west);
        \draw[->, very thick](classifier.east) -- node[rectangle, very thick, draw=black, fill=gray!20] {$N_2$} (maf2.west);
        \draw[->, very thick](classifier.east) -- node[rectangle, very thick, draw=black, fill=gray!20] {$N_3$}(maf3.west);
        \draw[->, very thick](classifier.east) -- node[rectangle, very thick, draw=black, fill=gray!20] {$N_4$} (maf4.west);

        \draw[<-, very thick](maf1.east) -- node[rectangle, very thick, draw=black, fill=gray!20] {$\mathcal{W}_1 = \sum_{j=0}^{N_1} w_{1, j}$} (output_1.west);
        \draw[<-, very thick](maf2.east) -- node[rectangle, very thick, draw=black, fill=gray!20] {$\mathcal{W}_2 = \sum_{j=0}^{N_2} w_{2, j}$} (output_1.west);
        \draw[<-, very thick](maf4.east) -- node[rectangle, very thick, draw=black, fill=gray!20] {$\mathcal{W}_3 = \sum_{j=0}^{N_3} w_{3, j}$} (output_1.west);
        \draw[<-, very thick](maf3.east) -- node[rectangle, very thick, draw=black, fill=gray!20] {$\mathcal{W}_4 = \sum_{j=0}^{N_4} w_{4, j}$} (output_1.west);
        
    \end{tikzpicture}
    \caption{The figure shows the training and sampling process for our piecewise NFs. From left to right, we start with a set of samples $x$ drawn from $p_X(x)$ and classify the samples into clusters (e.g. using $k$-means), determining the number of clusters to use based on the silhouette score. We then train a Masked Autoregressive Flow on each cluster $k$ using a standard normal distribution as our base. When drawing samples from $p_\theta(x)$ we select which MAF to draw from using weights defined according to the total cluster weight.}
    \label{fig:flow_diagram}
\end{figure*}
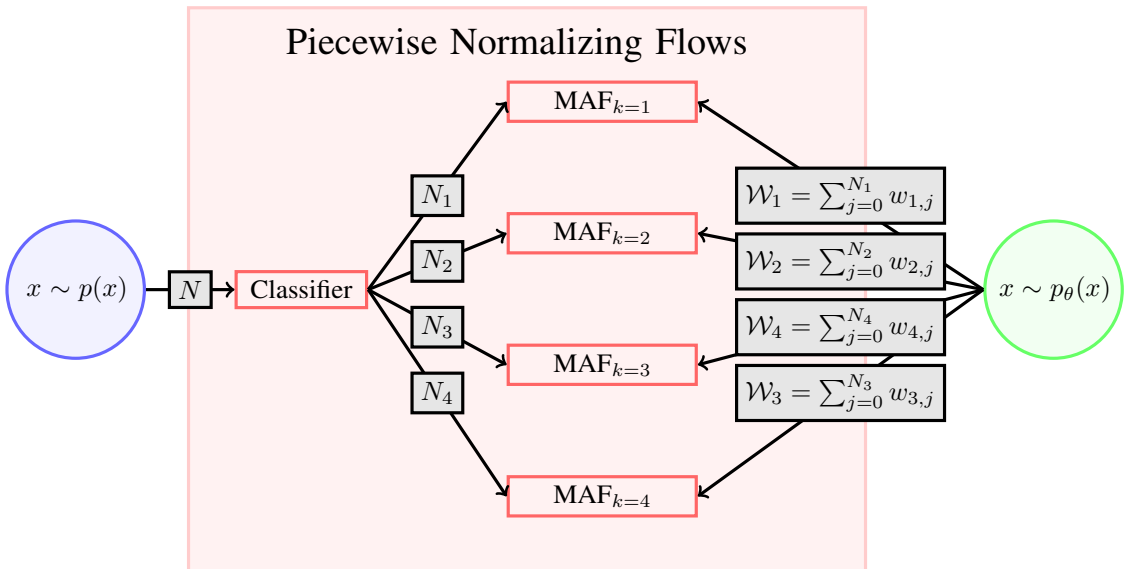

Our approach is independent of the choice of clustering algorithm (see \cref{sec:clustering}), however in this paper we use the $k$-means algorithm \citep{Hugo1957, MacQueen1967}.

We can draw samples from each cluster in a trained piecewise normalizing flow with a weight given by the total weight in the target cluster, $\mathcal{W}_k = \sum_{j=0}^{N_k} w_{j, k}$. This process is shown in \cref{fig:flow_diagram} along with the general flow of classification and training.

The probability of our piecewise MAF for a sample, $x$, is given by
%\begin{equation}
%    p_\theta (x) = \frac{1}{\sum_k \mathcal{W}_k} \sum_k \bigg(\mathcal{W}_k \exp \bigg( \log p_{\theta_k}(x)\bigg) \bigg)
%\end{equation}
%or in log-space
\begin{equation}
    \begin{aligned}
    \log p_\theta (x) = & \log \bigg(\sum_k \mathcal{W}_k \exp\bigg(\log p_{\theta_k}(x) \bigg)\bigg) \\ & - \log \bigg( \sum_k \mathcal{W}_k \bigg),
    \end{aligned}
\end{equation}
or equivalently
\begin{equation}
    \begin{aligned}
   \log p_\theta (x) = & \log\bigg( \sum_k \bigg[ \exp \bigg( \log (\mathcal{W}_k) + \\ & \log (p_{\theta,k})\bigg)\bigg]\bigg) - \log \bigg( \sum_k \mathcal{W}_k \bigg),
    \end{aligned}
\end{equation}
and if we normalize $\mathcal{W}_k$ such that the sum is equal to one we arrive at
\begin{equation}
    \log p_\theta (x) =  \log\bigg( \sum_k \bigg[ \exp \bigg( \log (\mathcal{W}_k) + \log (p_{\theta,k})\bigg)\bigg]\bigg).
\end{equation}
From this we can calculate the KL divergence and an associated Monte Carlo error 
\begin{equation}
    \begin{aligned}
    D_{KL} = \bigg\langle \log & \frac{p_\theta(x)}{p_X(x)}\bigg\rangle_{p_\theta(x)} \\ &\pm \frac{1}{\sqrt{N}} \sigma \bigg(\log\frac{p_\theta(x)}{p_X(x)}\bigg)_{p_\theta(x)},
    \end{aligned}
    \label{eq:kl}
\end{equation}
where $p_X(x)$ is the analytic target distribution, $N$ samples $x\sim p_\theta(x)$ and $\sigma(f)_g$ is the standard deviation of $f$ with respect to $g$.

\section{Benchmarking}
\label{sec:benchmarking}

\subsection{Toy Models}

\begin{table*}
\begin{center}
\begin{tabular}{|c|c|c|c|}
\hline
     Distribution & \multicolumn{3}{c|}{KL divergence} \\
     \cline{2-4}
     & MAF & RealNVP & Piecewise MAF \\
     & Gaussian Base & Resampled Base & Gaussian Base \\
     & e.g. \cite{MAFs} & \cite{Stimper2021} & This work \\
     \hline
     Two Moons & $-0.054\pm 0.009 $  & $0.028 \pm 0.055$  & $\mathbf{0.004 \pm 0.022}$ \\
     Circle of Gaussians & $ -0.046 \pm 0.050 $ & $ 0.231 \pm 0.138 $ & $\mathbf{0.001 \pm 0.005}$ \\
     Two Rings & $ 0.015 \pm 0.078 $ & $  0.080 \pm 0.106 $ & $ \mathbf{0.031 \pm 0.032}$ \\
     \hline
\end{tabular}
\end{center}
\caption{The table shows the KL divergence between the target samples for a series of multi-modal distributions and the equivalent representations with a single MAF, a RealNVP with a resampled base distribution \citep[e.g][]{Stimper2021} and a piecewise MAF. We train each type of flow on the target distributions ten times. The reported KL divergences are the mean over the ten training runs. The error for each training run is calculated as in \cref{eq:kl} and the errors are added in quadrature across training runs to get the error on the mean KL.
The distributions and an example set of samples drawn from each of the flows is shown in \cref{fig:benchmarks}. We have used early stopping for training and each flow has approximately the same number of hyperparameters. We find that our piecewise approach is competitive to the approach detailed in \cite{Stimper2021} for modelling multi-modal probability distributions and, since the KL divergence have a smaller error, more consistent over repeated training runs. Negative KL divergences are a result of the Monte Carlo nature of the calculation.}
\label{tab:benchmarks}
\end{table*}

\begin{figure*}
    \centering
    \includegraphics[width=0.7\linewidth]{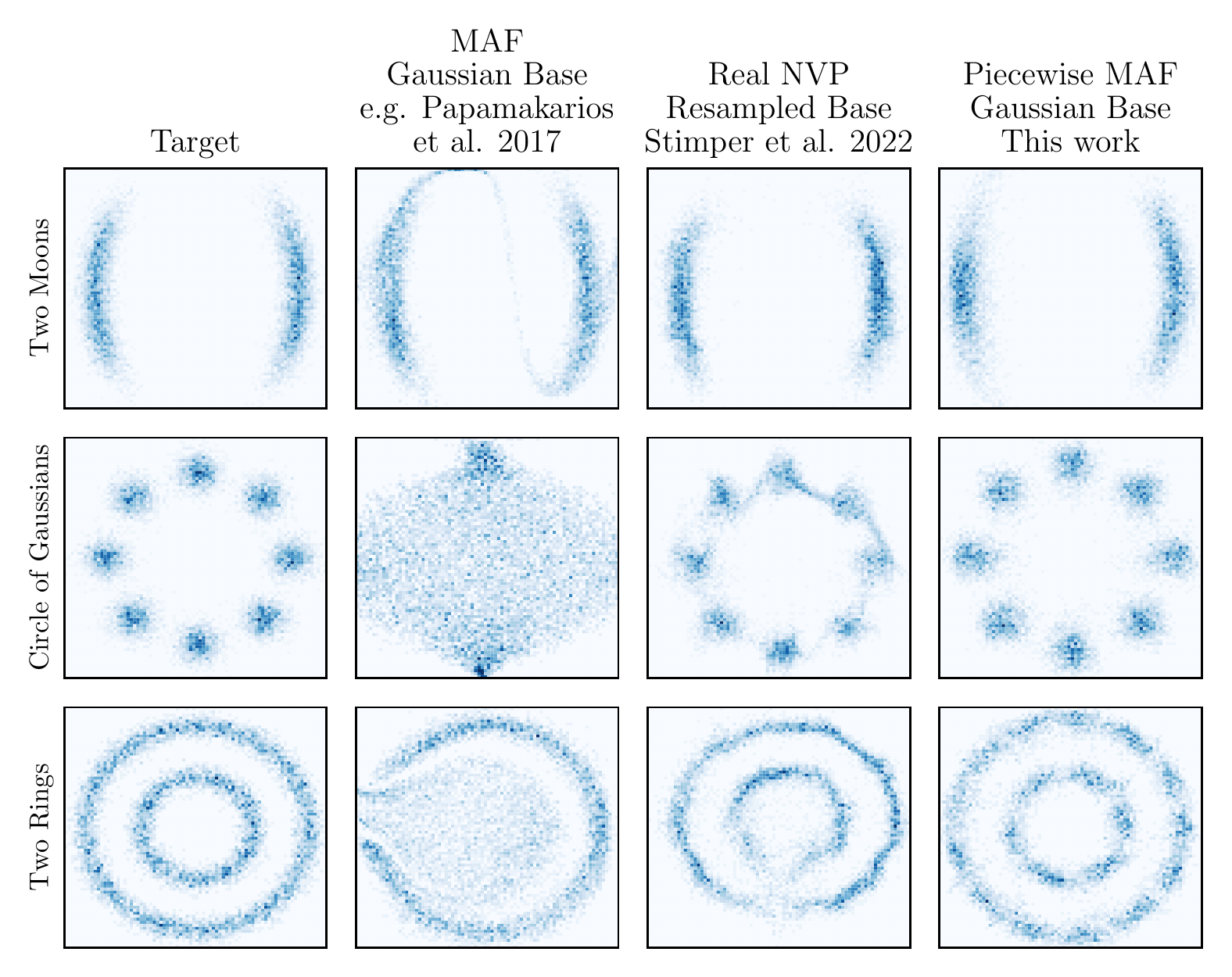}
    \caption{The graph shows a series of multi-modal target distributions as 2D histograms of samples in the first column and three representations of this distribution produced with three different types of normalizing flow. We use early stopping, as described in the main text, to train all the normalizing flows. The second column shows a simple Masked Autoregressive Flow with a Gaussian base distribution based on the work in \cite{MAFs} and the later two columns show two approaches to improve the accuracy of the model. The first is from \cite{Stimper2021} and uses a RealNVP flow to model the distributions while resampling the base distribution to better match the topology of the target distribution. In the last column, we show the results from our PNFs, where we have effectively modified the topology of the target distribution to better match the Gaussian base distribution by performing clustering.}
    \label{fig:benchmarks}
\end{figure*}

We test our piecewise normalizing flow on a series of multi-modal problems that are explored in \cite{Stimper2021}. For each, we model the distribution using a single MAF with a Gaussian base distribution, a RealNVP with a resampled base distribution as in \citet{Stimper2021} and our piecewise MAF approach. While there are a number of different methods for accurately learning multimodal distributions, the vast majority are not bijective unlike the approach taken in \citet{Stimper2021} and so we compare our results with that method. We calculate the Kullback-Leibler divergence between the original target distribution and the trained flows to assess the performance of the different methods.

We show the target distributions and samples drawn from each type of flow, a MAF, a piecewise MAF and a RealNVP with a resampled base distribution, in \cref{fig:benchmarks} and report the corresponding KL divergences in \cref{tab:benchmarks}. We train each flow ten times and calculate an average KL divergence for each flow and distribution. We add the associated Monte Carlo errors in quadrature to get an error on the mean KL divergence. We find that our piecewise approach outperforms the resampled base distribution approach of \citet{Stimper2021} for all of our multi-modal benchmarks. We use the code available at \url{https://github.com/VincentStimper/resampled-base-flows} 
to fit the RealNVP flows and the code %\textcolor{blue}{IDENTIFYING REFERENCE AND URL}
\textsc{margarine} \citep{Bevins2022a, Bevins2022b} available at \url{https://github.com/htjb/margarine} 
to fit the MAFs and PNFs. We chose the architecture of the MAF, PNF and RealNVP such that they have approximately the same number of hyperparameters.

We use early stopping to train all the flows in the paper, including the RealNVP resampled base models from \citet{Stimper2021}. For the single MAF and RealNVP, we use 20\% of the samples from our target distribution as test samples. We early stop when the minimum test loss has not improved for 2\% of the maximum iterations requested, 10,000 for all examples in this paper, and roll back to the optimum model. For the PNF, we split each cluster into training and test samples and perform early stopping on each cluster according to the same criterion, meaning that each cluster is trained for a different number of epochs. All the experiments in this paper were run on an Apple MacBook Pro with an M2 chip and 8GB memory.

In \cref{tab:benchmark_times}, we report the cost of training, the number of epochs needed and the time taken on average to train a MAF, RealNVP and a PNF on the benchmark distributions five times. 
We define the cost of training the MAF and RealNVP as the product of the number of epochs, $E$, the length of the training and test data combined $N$ and the number of hyperparameters in the network $h$
\begin{equation}
    \mathcal{C} = E N h,
\end{equation}
the idea being that at each epoch we do $h$ calculations $N$ times (since we evaluate both the test and training loss). For the PNF each piece of the distribution is trained on a fraction of the data, for a different number of epochs and each MAF in the PNF has $h_k < h$ such that $\sum_k h_k \approx h$. The cost of such a `divide and conquer` approach is given by
\begin{equation}
    \mathcal{C} = \sum_k E_k N_k h_k.
\end{equation}
We report the cost and associated errors for the RealNVP and PNF normalized to the value for the MAF for ease of comparison. We find that the RealNVP requires the fewest number of epochs and has the smallest wall clock time. However, the reported wall clock time for the PNF corresponds to training each piece in series and can be made more competitive with the RealNVP through parallelisation. This is represented in the significantly lower costs for PNF training.

\begin{table*}
    \begin{center}
\begin{tabular}{|c|c|c|c|}
\hline
    \multicolumn{4}{|c|}{Cost} \\
    \hline
     Distribution & MAF & RealNVP & PNF \\
     \hline
     Two Moons & $1.000 \pm 0.191 $ & $0.241 \pm 0.003 $ & $\mathbf{0.225 \pm 0.037}$ \\
     Circle of Gaussians & $1.000 \pm 0.206 $  & $0.794 \pm 0.092$ & $\mathbf{0.226 \pm 0.005}$\\
     Two Rings & $1.000 \pm 0.265 $ & $0.800 \pm 0.098$ & $\mathbf{0.236 \pm 0.007}$ \\
     \hline
\end{tabular}\\
\begin{tabular}{|c|c|c|c|}
\hline
    \multicolumn{4}{|c|}{Epochs} \\
    \hline
     Distribution & MAF & RealNVP & PNF \\
     \hline
     Two Moons &  $3131 \pm 598$  & $\mathbf{655 \pm 9}$ & $2118 \pm 351$\\
     Circle of Gaussians & $1765 \pm 364$ & $\mathbf{1217 \pm 141}$ & $19130 \pm 412$ \\
     Two Rings &  $1896 \pm 503$ & $\mathbf{1317 \pm 162}$ & $42444 \pm 1720$\\
     \hline
\end{tabular}\\
\begin{tabular}{|c|c|c|c|}
\hline
    \multicolumn{4}{|c|}{Wall Time [s]} \\
    \hline
     Distribution & MAF & RealNVP & PNF \\
     \hline
     Two Moons & $72.80 \pm 12.72$ & $\mathbf{4.05 \pm 0.05}$ & $22.72 \pm 3.31$ \\
     Circle of Gaussians & $42.13 \pm 8.49$ & $\mathbf{7.62 \pm 0.86}$& $26.47 \pm 1.15$\\
     Two Rings & $46.02 \pm 12.14$& $\mathbf{8.33 \pm 1.04}$& $44.55 \pm 1.82$\\
     \hline
\end{tabular}
\end{center}
\caption{The table summarises the cost of training, the total number of epochs and the wall time for a MAF, a RealNVP and a PNF on the three toy example distributions explored in this paper. The values are averages over five training runs with an associated Monte Carlo error. Here the cost, as defined in the text, and the associated error are normalized to the MAF value to aid the comparison. Of the three tested algorithms the RealNVP requires the fewest epochs to train and has the shortest wall time. However, the cost to train the PNF is significantly lower than the RealNVP and MAF because the data are distributed over several smaller MAFs with fewer hyperparameters. It should be noted that the total wall clock time for the PNF corresponds to training the pieces in series and through parallelisation the PNF wall clock time can be made competitive with the RealNVP. This is represented in the cost.}
\label{tab:benchmark_times}
\end{table*}

\subsection{Choosing a clustering algorithm}
\label{sec:clustering}

Next, we illustrate how the choice of clustering algorithm can impact the accuracy of the PNF. We repeat the analysis from the previous section on the two rings distribution, using five different clustering algorithms to perform the initial division of the target samples. 

So far, we have employed the $k$-means clustering algorithm to identify clusters in the target distribution. %$k$-means works by adjusting the position of $k$ centroids such that the sum of squares between the cluster data points and centroids is minimised. 
Here we test Mini-batch $k$-means \citep{minibatch_kmeans}, the Mean Shift algorithm \cite{Cheng1995}, Spectral Clustering \cite{Shi2000}, Agglomerative Clustering \cite{Nielsen2016} and the Birch algorithm \cite{Zhang1996}. All of these algorithms are readily accessible through the \textsc{sklearn} \textsc{python} package which we used with the default settings, and we refer the reader to the review by \citet{saxena2017} for more details.

For each clustering algorithm, we run training ten times and calculate a KL divergence between the target distribution and the PNF according to \cref{eq:kl}. The mean KL divergence values over training runs and an associated error are reported in \cref{tab:clustering_algorithm}. We can clearly see that the $k$-means algorithm outperforms the others but that Mean Shift and Birch clustering competitive with each other, $k$-means and the RealNVP resampled base approach.

This work is not intended to be an exhaustive search of available clustering algorithms, however, it illustrates that PNFs can be built with different clustering algorithms. We recommend $k$-means, Mean Shift and Birch clustering based on the results presented here but stress that the choice of clustering algorithm may be data dependent.

%It is also important to note here that because we restricted the number of networks in the PNFs so that the number of hyperparameters are approximately consistent across the different tested flows, for some of the clustering algorithms there is only one network per cluster and as such the flow is not as expressive as one would like. This is believed to lead to the poor performance of the Spectral Clustering, for example. In practice, the number of networks and the network architecture needs to be optimized.

\begin{figure*}
    \centering
    \includegraphics[width=0.7\linewidth]{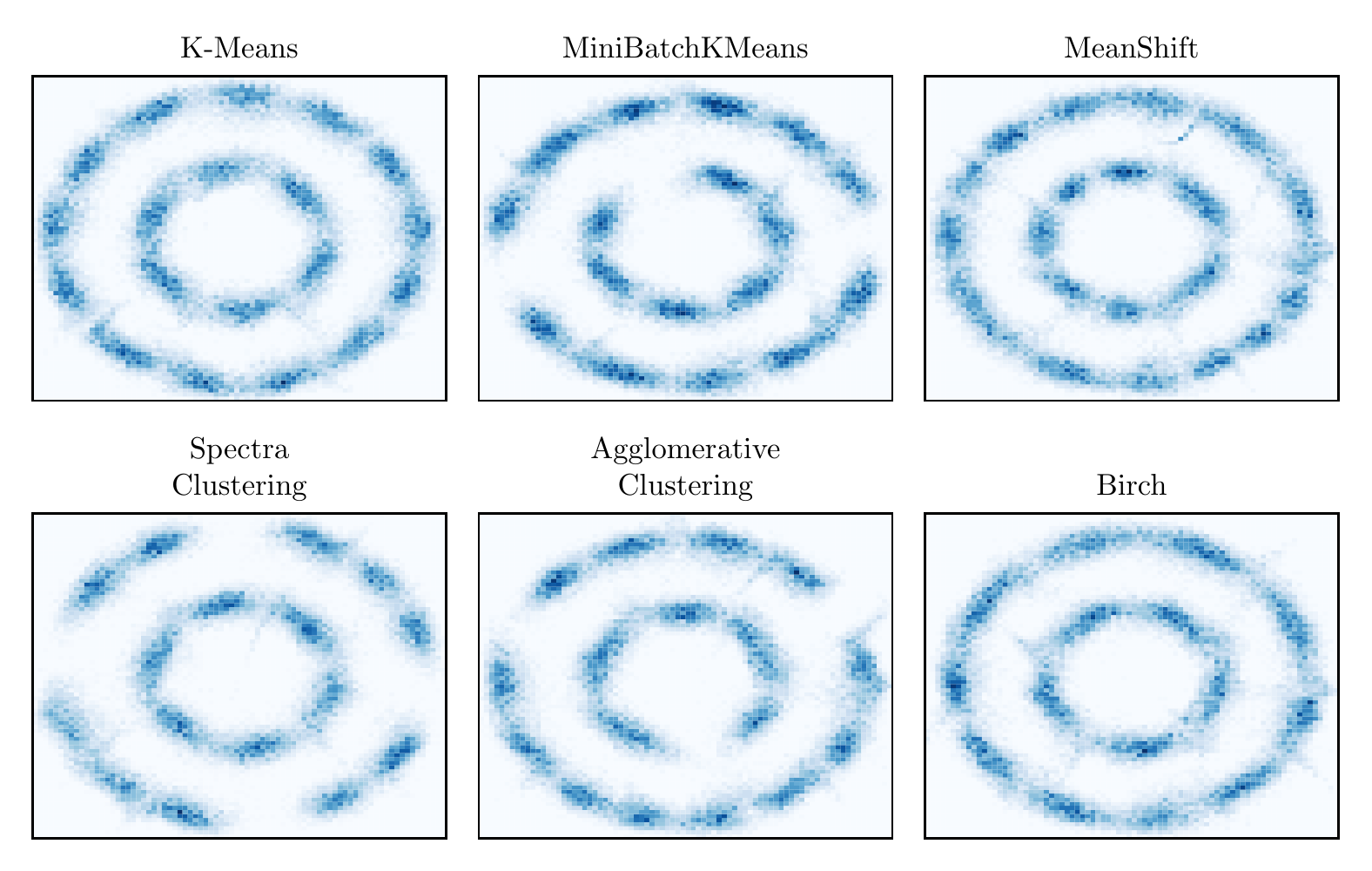}
    \caption{The figure shows samples drawn from various different PNFs built using different clustering algorithms, along with samples from the true distribution and samples from a single masked autoregressive flow. The figure illustrates that the PNF can be effectively built from using different clustering algorithms and, along with \cref{tab:clustering_algorithm}, that some clustering algorithms perform better than others. We choose network architectures for our PNFs and MAF such that they all have approximately the same number of hyperparameters.}
    \label{fig:clustering_figure}
\end{figure*}

\begin{table}[]
    \centering
    \begin{tabular}{|c|c|}
        \hline
         Flow & $D_{KL}$ \\
         Type  &  \\
         \hline
         MAF & $ 0.015 \pm 0.078 $  \\
         RealNVP Resampled Base & $  0.080 \pm 0.106 $\\
         PNF $k$-means & $ \mathbf{0.031 \pm 0.032}$\\
          PNF Mini-batch $k$-means & $0.847 \pm 0.038$ \\
         PNF Mean Shift & $0.040 \pm 0.007$\\
         PNF Spectral Clustering & $0.814 \pm 0.035$\\
         PNF Agglomerative Clustering & $0.419 \pm 0.028$\\
         PNF Birch & $0.036 \pm 0.007$\\
         \hline
    \end{tabular}
    \caption{The table shows the KL divergence between the target distribution, the two rings toy example, and a series of different PNFs built with different clustering algorithms, a single MAF and the RealNVP with a resampled base distribution. The values for the MAF, RealNVP and $k$-means PNF are drawn from \cref{tab:benchmarks}. We find that $k$-means is the best of the tested clustering algorithms, but that Mean Shift and Birch Clustering are competitive with the RealNVP.}
    \label{tab:clustering_algorithm}
\end{table}

\begin{table}[t!]
    \centering
    \begin{tabular}{|c|c|c|}
    \hline
         Flow Type &  POWER (6) & GAS (8)  \\
         \hline
         MAF & $\mathbf{-0.05 \pm 0.08}$ & $\mathbf{5.10 \pm 0.09}$  \\
         PNF & $\mathbf{-0.09 \pm 0.07}$ & $\mathbf{4.32 \pm 0.68}$ \\
         \hline
    \end{tabular}
    \begin{tabular}{|c|c|c|}
    \hline
         Flow Type &   HEPMASS (21) & MINIBOONE (43) \\
         \hline
         MAF &  $\mathbf{-20.11 \pm 0.02}$ & $\mathbf{-11.63 \pm 0.44}$ \\
         PNF & $-20.71 \pm 0.02$ & $-12.40 \pm 0.45 $ \\
         \hline
    \end{tabular}
    \caption{We compare the accuracy of the PNF proposed here with a single MAF on four data sets from the UCI machine learning repository \cite{uci}. The table shows the average log-likelihood evaluated over a test data set for the POWER, GAS, HEPMASS and MINIBOONE UCI data sets. We show the number of dimensions for each data set in brackets. Generally speaking, the MAF performs better than the PNF however \citet{MAFs} previously found that generally the data sets could be better modelled with a MAF with a standard normal base distribution over a Gaussian mixture base distribution suggesting that the data sets are not significantly multi-modal. Again we use early stopping during training and note that the PNF and MAF have approximately the same number of hyperparameters. Errors are two sigma as in \citet{MAFs}.}
    \label{tab:physical_benchmarks}
\end{table}

\subsection{Physical Examples}

It is common to compare proposed normalizing flow architectures on data sets from the UCI machine learning repository \cite{uci}. Accuracy of emulation is assessed by evaluating the average log-probability over the flow for a set of unseen test samples. A higher average log-probability indicates a more accurate network, which can be seen from the discussion of the loss function in \cref{sec:normalizing_flows}. We report the results of training on four UCI datasets (POWER, GAS, HEPMASS and MINIBOONE) in \cref{tab:physical_benchmarks} along with an associated Monte Carlo error bar.

The POWER and GAS data sets are much larger than HEPMASS and MINIBOONE and consequently computationally expensive to train on. We therefore took representative subsets of these two data sets for training and testing.

Generally, speaking the MAF performs better on these physical data sets when compared to the PNF although the PNF is competitive for the POWER and GAS data sets. We note that there is no indication in the literature that these data sets are multi-modal. For example, \citet{MAFs} find that generally MAFs with standard normal base distributions perform better on these targets compared to MAFs with Gaussian Mixture model base distributions. We therefore do not necessarily expect the PNFs to perform better than the MAFs, as was shown for the toy multi-modal cases. Indeed, for all the data sets we see a preference via the silhouette score for the minimum tested number of clusters, 2.
It is reassuring, however, that the PNF does not perform drastically worse than the MAF on these examples.

\section{Limitations}

%The bijective and differential properites of normalizing flows allow you 
Normalizing flows are useful because they are bijective and differentiable, however these properties are sometimes lost when trying to tackle multi-modal distributions. Our piecewise approach is differentiable, provided that the optimization algorithm used to classify the samples from the target distribution into clusters is also differentiable and $k$, the number of clusters, is fixed during training. %Differentiability is lost if $k$ changes, however, one can imagine an improved $k$ picking algorithm where $k$ is determined through minimisation of the loss in \cref{eq:loss} rather than via a silhouette score. This is left for future work.

Our piecewise approach is bijective if we know the cluster that our sample has been drawn from. In other words, there is a one to one mapping between a set of real numbers drawn from an $N$ dimensional base distribution to a set of real numbers on the $N$ dimensional target distribution plus a cluster number, $n\in \mathbb{N}$
\begin{equation}
    \mathbb{R}^N_{\mathcal{N}(\mu = 0, \sigma=1)}\Longleftrightarrow \mathbb{R}^N_{p_\theta(x)} \times \mathbb{N}.
\end{equation}
%Again with an improved $k$ picking algorithm the the dependence on the cluster number could be relinquished.

%Our approach is also limited by the complexity of the target distribution and the number of clusters required to describe the distribution. However, this issue is not unique to the PNF and the number of required clusters is an issue of computational power which can be overcome by training each NF in parallel.

\section{Conclusions}

In this work, we demonstrate a new method for accurately learning multi-modal probability distributions with normalizing flows through the aid of an initial classification of the target distribution into clusters.

Normalizing flows generally struggle with multi-modal distributions because there is a significant difference in the topology of the base distribution and the target distribution to be learnt. By dividing our target distribution up into clusters, we are essentially, for each piece, making the topology of the target distribution more like the topology of our base distribution, leading to an improved expressivity. \citet{Stimper2021} by contrast, resampled the base distribution so that it and the target distribution shared a more similar topology. We find that both approaches perform well on our benchmark tests, but that our piecewise approach is better able to capture the distributions when training with early stopping. The piecewise nature of our approach allows the MAFs to be trained in parallel. %Each MAF in our network is bijective and provided the clustering algorithm is differentiable our piecewise approach is also differentiable.

Through the combination of clustering and normalizing flows, we are able to produce accurate representations of complex multi-modal probability distributions.

\section{Acknowledgements}

HTJB acknowledges support from the Kavli Institute for Cosmology, Cambridge, the Kavli Foundation and of St Edmunds College, Cambridge.  WJH thanks the Royal Society for their support through a University Research Fellowship. TGJ would like to thank the Science and Technology Facilities Council (UK) for their continued support through grant number ST/V506606/1.

This work used the DiRAC Data Intensive service (CSD3) at the University of Cambridge, managed by the University of Cambridge University Information Services on behalf of the STFC DiRAC HPC Facility (www.dirac.ac.uk). The DiRAC component of CSD3 at Cambridge was funded by BEIS, UKRI and STFC capital funding and STFC operations grants. DiRAC is part of the UKRI Digital Research Infrastructure.

\section{Impact Statement}

Bias in machine learning is currently an active area of concern and research \citep[e.g. ][]{jiang2020identifying, Mehrabi2021}. While we have focused on normalizing flows for probability density estimation in this work, we believe that the general approach has wider applications. One can imagine dividing a training data set into different classes and training a series of generative models to produce new data. One could then draw samples based on user defined weights rather than data defined weights, as is done here. As such, one can begin to overcome some selection biases when training machine learning models. Of course the `user determined' nature of the weights could be used to introduce more bias into the modelling, therefore the method through which they are derived would have to be explicit and ethically sound.

\bibliography{ref}
\bibliographystyle{icml2024}

%%%%%%%%%%%%%%%%%%%%%%%%%%%%%%%%%%%%%%%%%%%%%%%%%%%%%%%%%%%%%%%%%%%%%%%%%%%%%%%
%%%%%%%%%%%%%%%%%%%%%%%%%%%%%%%%%%%%%%%%%%%%%%%%%%%%%%%%%%%%%%%%%%%%%%%%%%%%%%%
% APPENDIX
%%%%%%%%%%%%%%%%%%%%%%%%%%%%%%%%%%%%%%%%%%%%%%%%%%%%%%%%%%%%%%%%%%%%%%%%%%%%%%%
%%%%%%%%%%%%%%%%%%%%%%%%%%%%%%%%%%%%%%%%%%%%%%%%%%%%%%%%%%%%%%%%%%%%%%%%%%%%%%%
%\newpage
%\appendix
%\onecolumn
%\section{You \emph{can} have an appendix here.}

%You can have as much text here as you want. The main body must be at most $8$ pages long.
%For the final version, one more page can be added.
%If you want, you can use an appendix like this one, even using the one-column format.
%%%%%%%%%%%%%%%%%%%%%%%%%%%%%%%%%%%%%%%%%%%%%%%%%%%%%%%%%%%%%%%%%%%%%%%%%%%%%%%
%%%%%%%%%%%%%%%%%%%%%%%%%%%%%%%%%%%%%%%%%%%%%%%%%%%%%%%%%%%%%%%%%%%%%%%%%%%%%%%

\end{document}